\documentclass{article}
\usepackage[utf8]{inputenc}
\usepackage[table,xcdraw]{xcolor}
\usepackage{authblk}
\usepackage{graphicx}
\usepackage[a4paper, total={6in, 8in}]{geometry}

\title{Realistic Video Summarization through VISIOCITY: A New Benchmark and Evaluation Framework}

\author[1]{Vishal Kaushal}
\author[2]{Suraj Kothawade}
\author[2]{Rishabh Iyer}
\author[1]{Ganesh Ramakrishnan}
\affil[1]{Department of Computer Science, Indian Institute of Technology Bombay}
\affil[2]{Department of Computer Science, University of Texas at Dallas}

\begin{document}

\maketitle

\begin{abstract}
  Automatic video summarization has attracted a lot of interest. However it is still an unsolved problem due to several challenges. We take steps towards making automatic video summarization more realistic by addressing the following challenges. Firstly, the currently available datasets either have very short videos or have few long videos of only a particular type. We introduce a new benchmarking dataset called VISIOCITY which comprises of longer videos across six different categories with dense concept annotations capable of supporting different flavors of video summarization and can lend itself well to other vision problems. Secondly, for long videos, human reference summaries, necessary for supervised video summarization techniques, are difficult to obtain. We present a novel recipe based on pareto optimality to automatically generate multiple reference summaries from indirect ground truth present in VISIOCITY. We show that these summaries are at par with human summaries. Thirdly, we demonstrate that in the presence of multiple ground truth summaries (due to the highly subjective nature of the task), learning from a single combined ground truth summary using a single loss function is not a good idea. We propose a simple recipe to enhance an existing model using a combination of losses and demonstrate that it beats the current state of the art techniques when tested on VISIOCITY. Towards this we present a study of different characteristics of a summary and demonstrate how, especially in long videos, it is quite possible and frequent to have two good summaries with different characteristics. Consequently, a single measure to evaluate a summary, as is the current typical practice, falls short. We propose a framework for better quantitative assessment of summary quality which is closer to human judgment than a single measure, say F1. We report the performance of a few representative techniques of video summarization on VISIOCITY assessed using various measures and bring out the limitation of the techniques and/or the assessment mechanism in modeling human judgment and demonstrate the effectiveness of our evaluation framework in doing so.\looseness-1
\end{abstract}

%%\keywords{video summarization, benchmark, dataset, automatic evaluation, performance}

\section{Introduction}

Videos have become an indispensable medium for capturing and conveying information in many sectors like entertainment (TV shows, movies, etc.), sports, personal events (birthday, wedding etc.), education (HOWTOs, tech talks etc.), to name a few. The increasing availability of cheaper and better video capturing and video storage devices have led to the unprecedented growth in the amount of video data available today. Most of this data, however, is plagued with redundancy, partly because of the inherent nature of videos (as a set of {\it many} images) and partly due to the 'capture-now-process-later' mentality. Consequently this has given rise to the need of automatic video summarization techniques which essentially aim at producing shorter videos without significantly compromising the quality and quantity of information contained in them. What makes automatic video summarization especially challenging though is that an {\it ideal} summary is highly context dependent (depends on the purpose behind getting a video summary), subjective (even for the same purpose, preferences of two persons don't match) and depends on high-level semantics of the video (two visually different scenes could capture the same semantics or visually similar looking scenes could capture different semantics). Thus, there is no single right answer. For example, as far as context is concerned, one may want to summarize a surveillance video either to see a 'gist' of what all happened or to quickly spot any 'abnormal' activity. Similarly, as far as personal preferences or subjectivity is concerned, while summarizing a {\it Friends} video (a popular TV series), two users may have different opinion on what is 'important' or 'interesting'. As another example, when it comes to semantics at a higher level than what is captured by visual appearance, closeup of a player in soccer can be considered important if it is immediately followed by a goal, while not so important when it occurs elsewhere. These and other issues discussed below make it an interesting research problem with several papers pushing the state-of-the-art for newer algorithms and model architectures ~\cite{fu2019attentive, yuan2019cycle, zhang2016video, zhang2016summary, zhou2018deep, xiong2019less, gygli2015video, vasudevan2017query, li2018local} and datasets~\cite{GygliECCV14, potapov2014category, conf/cvpr/SongVSJ15}. However, as noted by a few recent works several fundamental issues remain to be addressed.  We summarize these below.

\textbf{Dataset:} For a true comparison between different techniques, a benchmark dataset is critical. Almost all recent techniques have reported their results on TVSum \cite{song2015tvsum} and SumMe \cite{gygli2014creating} which have emerged as benchmarking datasets of sorts. However, since the average video length in these datasets is of the order of \emph{only} 1-5 minutes, they are far from being effective in real-world settings. The effect and need of automatic video summarization is desired and more pronounced especially in long videos. While there have been many  attempts at creating datasets for video summarization, they either a) have very short videos, or b) they have very few long videos and often only a particular type. A large dataset with a lot of different types of full-length videos with rich annotations to be able to support different techniques was one of the recommendations in \cite{truong2007video}, is still not a reality and is clearly a need of the hour \cite{ji2019video}. One of the contributions of this work is to address this gap. We present a large and diverse video summarization dataset VISIOCITY (Video Summarization Based on Continuity, Intent and Diversity) with a total of 67 videos spanning six categories: TV shows (\emph{Friends}) , sports (soccer), surveillance, education (tech-talks), birthday videos and wedding videos. The videos have an average duration of about 50 minutes. In Table 1 and Table 2 we provide a comparison of VISIOCITY with other existing datasets. Furthermore, different flavors of video summarization, for example, query focused video summarization \cite{xiao2020convolutional, vasudevan2017query, sharghi2017query}, are often treated differently and on different datasets. With its rich annotations (described in section~\ref{subsec3:annotations}), VISIOCITY lends itself well to other flavors of video summarization and also other computer vision video analysis tasks like captioning or action detection. Also, since the videos span across different well-defined categories, VISIOCITY is also suitable for more in-depth domain specific studies on video summarization \cite{truong2007video, potapov2014category}

\textbf{Nature of supervision:} Supervised techniques tend to work better than unsupervised techniques \cite{ji2019video, zhang2016video} because of learning directly from user summaries. In a race to achieve better performance, most state of the art techniques are based on deep architectures and are thus data hungry. The larger the dataset and more the number of reference summaries to learn from, the better. However, it is very costly to create ground truth summaries when it comes to long videos. It becomes increasingly expensive and, beyond a point, infeasible to get these reference summaries from humans. Also, this is not scalable to experiments where reference summaries of different lengths are desired \cite{gygli2015video}. We address this need by providing a recipe for ground truth creation from indirect ground truth provided by humans. Indirect ground truth is annotations in the form of concepts present in each shot. We argue, getting indirect ground truth for long videos is more feasible and much more objective. As elaborated in section~\ref{subsec4:autosumm}, this allows for automatically generating as many different ground truth summaries as desired by a machine learning model and more importantly, with different characteristics. Since there is no single right answer, indirect ground truth annotations, being at a meta level, can be seen as a \emph{generator} of ground truth summaries \cite{truong2007video}.\looseness-1

A related problem is the fact that current supervised techniques are trained using a 'combined' ground truth summary, either in form of combined scores from multiple ground truth summaries or scores \cite{zhang2016summary,fajtl2018summarizing, ji2019video} or in form a set of ground truth selections, as in dPPLSTM \cite{zhang2016summary}. However, since there can be multiple correct answers, a reason for low consistency between user summaries ~\cite{kannappan2019human, otani2019rethinking} combining them into one misses out on the separate flavors captured by each of them. Combining many into one set of scores also runs the risk of giving more emphasis to 'importance' over and above other desirable characteristics of a summary like continuity, diversity etc. This is also noted by \cite{apostolidis2020unsupervised, zhou2018deep} where they argue that supervised learning approaches, which rely on the use of a combined ground-truth summary, cannot fully explore the learning potential of such architectures. The necessity to deal with different kind of summaries in different ways was also observed by \cite{truong2007video}. \cite{apostolidis2020unsupervised, zhou2018deep} use this argument to advocate the use of unsupervised approaches. Another alternative however is to make a model learn directly from multiple ground truth summaries instead of a 'combined' ground truth summary. The learning can be further enhanced by employing more than one or a combination of loss functions each measuring the deviation from different desired characteristics of good summaries. Using this principle, we demonstrate a simple extension to a mixture model (see section~\ref{sec5:propmodel}) which performs better than other state of the art techniques.

\textbf{Evaluation:} With a desire to be comparable across techniques, almost all recent work evaluates their results using F1 score defined as harmonic mean of precision (ratio of temporal overlap between candidate and reference summary to duration of summary) and recall (ratio of temporal overlap between candidate and reference summary to video duration) \cite{zhou2014learning, fajtl2018summarizing, ji2019video}. This approach of assessing a candidate summary vis-a-vis a ground truth summary sounds good, but it has following limitations: 1) The user summaries are themselves inconsistent with each other, as already noted above ~\cite{kannappan2019human, otani2019rethinking}. As a workaround, the assessment is done with respect to the nearest neighbor \cite{song2015tvsum, gygli2015video}. This is still not free from the problem of having limited ground truth summaries. The number of right answers can be large, especially in case of long videos and a good candidate may get a low score just because it was not fortunate to have a matching user summary. Furthermore, F1 seems to be good to measure the 'closeness' with a user summary, but the numbers can be deceptive as it is affected by the segmentation used as a post processing step in typical video summarization pipeline \cite{otani2019rethinking}. Another problem with F1 is that it is not well suited to measure other characteristics of a summary like continuity or diversity. Two summaries may have same F1 score, and yet one may be more continuous (and hence visually more pleasurable to watch) than another (more discussion on this in section~\ref{subsec4:goodsumm}). While F1 has its utility in measuring the goodness of a summary, instead of over dependence on one measure, we propose using a suite of measures to capture various aspects of a summary like continuity, diversity, redundancy, importance etc. As discussed in section~\ref{subsec4:goodsumm}, these use the annotation provided in VISIOCITY (indirect ground truth), as against comparing the candidate to ground truth summaries. 

In what follows, we first talk about the related work in the areas outlined above. Then we give details of the VISIOCITY benchmark dataset (section~\ref{sec3:visiocity}). In section~\ref{sec4:evalframework} we introduce different supervised scoring functions that can be used to characterize a good summary followed by a recipe to generate ground truth summaries and the proposed evaluation framework. Thereafter, in section~\ref{sec5:propmodel} we introduce a simple recipe to enhance a model which beats the state of the art by making use of multiple ground truth summaries and by learning using the multiple loss functions. Finally, in section~\ref{sec6:expresults} we present the extensive experiments and analysis demonstrating the different characteristics of good summaries, the effectiveness of our recipe, necessity of proposed evaluation framework, a better approach to learning and an analysis of the performance of a few representative state of the art techniques on our dataset. This is followed by conclusion and future work.\looseness-1 
\begin{table*}
\small{
\begin{center}
\begin{tabular}{ | c | c | c | c | c | c |}
\hline
Name & \# Videos & Duration of Videos & Total Duration & \# summ & \# cat\\
 \hline
  SumMe~\cite{GygliECCV14} & 25 & Avg: 2 min, 9 & 1 Hour, 10 min & 15-18 & - \\
 TVSum~\cite{conf/cvpr/SongVSJ15} & 50 & Avg: 4min, 11 sec & 3.5 Hours & 20 & 10\\
 MED Summaries~\cite{potapov2014category} & 260 & Dur: 1-5 mins, Avg: 2.5min & 9 Hours & 2-4 & 15 \\
 UT Egocentric~\cite{lee2012discovering} & 4 & Avg: 254 mins & 16 Hours & - & 1 \\
 Youtube 1~\cite{de2011vsumm} & 50 & Dur: 1-10 min, Avg: 1 min, 39 sec & 1.38 Hours & 5 & -\\
 Youtube 2~\cite{de2011vsumm} & 50 & Dur: 1-4 min, Avg: 2min, 54sec & 2.41 Hours & 5 & -\\
 Tour20~\cite{panda2017diversity} & 140 & Avg: 3 min & 7 Hours & - & - \\
 TV Episodes~\cite{yeung2014videoset} & 4 & Avg: 45 min & 3 Hours & - & 1 \\
 LOL~\cite{fu2017video} & 218 & Dur: 30 to 50 min & - & - & 1\\ 
 \textbf{VISIOCITY \ (OURS)} & \textbf{89} & \textbf{Dur: 14-121 mins, Avg: 55 mins} & \textbf{71 hours} & \textbf{160} & \textbf{5}\\
 \hline
 \end{tabular}
 \label{tab1:datasets}
\caption{Key statistics of the different video summarization datasets in literature. "-" means the corresponding information was not available. \# summ here means \# reference summaries per video. "Dur" stands for Duration and \#cat is \# categories or domains.}
 \end{center}}
\end{table*}

\section{Related Work} \label{sec2:related_work}

\subsection{Datasets} \label{subsec2:datasets}
As noted earlier, the currently available annotated datasets for video summarization have either too few videos or too short videos or videos of only one type. Table 1 summarizes the important statistics of the different datasets. MED Summaries Dataset~\cite{potapov2014category} consists of 160 annotated videos of length 1-5 minutes, with 15 event categories like birthday, wedding, feeding etc.  The annotation comprises of segments and their importance scores. TVSum~\cite{conf/cvpr/SongVSJ15} consists of 50 videos (average length 4 minutes) from 10 categories. with importance scores provided by 20 annotators for each 2 second snippet. The videos correspond to very short events like `changing vehicle tires`, `making sandwich` etc. The UT Egocentric Dataset~\cite{lee2012discovering}  consists of long and annotated videos captured from head mounted cameras. However, though each video is very long, there are only 4 videos and they are of one type, i.e. egocentric. SumMe~\cite{GygliECCV14} consists of 25 videos with an average length of about 2 min. The annotation is in form of user summaries of length between 5\% to 15\%. Each video has 15 summaries. The VSUMM dataset~\cite{de2011vsumm} consists of two datasets. Youtube consists of 50 videos (1-10 min) and OVP consists of 50 videos of about 1-4 min from the Open Video project. Each video has 5 user summaries in the form of set of key frames. Tour20~\cite{panda2017diversity} consists of 140 videos with a total duration of 7 hours and is designed primarily for multi video summarization. It is a collection of videos of a tourist place. The average duration of each video is about 3 mins. TV Episodes dataset~\cite{yeung2014videoset} consists of 4 TV show videos, each of 45 mins. The total duration is 3 hours. A recent dataset, UGSum52 \cite{lei2019framerank} offers 52 videos with 25 user generated summaries each. LOL~\cite{fu2017video} consists of online eSports videos from the League of Legends. It consists of 218 videos with each video being between 30-50 mins. The associated summary videos are 5 - 7 mins long. While this dataset is significantly larger compared to the other datasets, it is limited only to a single domain, i.e. eSports. \cite{sharghi2018improving} have extended the UTE dataset to 12 videos and have provided concept annotations, but they are limited to only egocentric videos and do not support any concept hierarchy. The scores annotations, as in TVSum etc. are superior indirect ground truth annotations, but are limited only to importance scores. VISIOCITY on the other hand comes with dense concept annotations for each shot. To the best of our knowledge, VISIOCITY is one of its kind large dataset with many long videos spanning across multiple categories and annotated with rich concept annotations for each shot. 

\subsection{Techniques for Automatic Video Summarization}
\label{subsec2:algos}
A number of techniques have been proposed to further the state of the art in automatic video summarization. Most video summarization algorithms try to optimize several criteria such as diversity, coverage, importance and representation. \cite{gygli2015video,kaushal2019framework} proposed a submodular optimization framework with combining weighted terms for each of these criteria in a mixture and trained using loss augmented inference. Several recent papers have used deep learning to model importance. \cite{zhang2016video} propose a LSTM based model which they call vsLSTM. They also propose to add diversity using a Determinantal Point Process along with an LSTM (which they call dppLSTM). \cite{zhou2018deep} uses deep reinforcement learning and model with diversity and representation based rewards. \cite{ji2019video} uses an encoder-decoder based attention model and has obtained the best results on TV-Sum and Summe datasets. \cite{vasudevan2017query} studied query based summarization where they incorporate a query relevance model along with their submodular framework with diversity, representation and importance. \cite{fu2019attentive} attempts to address video summarization via attention-aware and adversarial training.

\subsection{Evaluation}
\label{subsec2:eval}

As presented above, evaluation is challenging task owing to the multiple definitions of success. Early approaches \cite{lu2013story, ma2002user} involved user studies with the obvious problem of cost and reproducibility. With a move to automativ evaluation, every new technique of video summarization came with its own evaluation criteria making it difficult to compare results different techniques. Some of the early approaches included VIPER \cite{doermann2000tools}, which addresses the problem by defining a specific ground truth format which makes it easy to evaluate a candidate summary, and SUPERSEIV \cite{huang2004automatic} which is an unsupervised technique to evaluate video summarization algorithms that perform frame ranking. VERT \cite{li2010vert} on the other hand was inspired by BLEU in machine translation and ROUGE in text summarization. Other techniques include pixel-level distance between keyframes~\cite{khosla2013large}, objects of interest as an indicator of similarity~\cite{lee2012discovering} and precision-recall scores over key-frames selected by human annotators~\cite{gong2014diverse}. It is not surprising thus that \cite{truong2007video} observed that researchers should at least reach a consensus on what are the best procedures and metrics for evaluating video abstracts. They concluded, that a detailed research that focuses exclusively on the evaluation of existing techniques would also be a valuable addition to the field. This is one of the aims of this work. More recently, computing overlap between reference and generated summaries has become the standard framework for video summary evaluation. However, all these methods which require comparison with ground truth summaries suffer from the challenges highlghted above. Yeung et al. observed that visual (dis)similarity need not mean semantic (dis)similarity and hence proposed a text based approach of evaluation called VideoSet. The candidate summary is converted to text and its similarity is computed with a ground truth textual summary. That text is better equipped at capturing higher level semantics has been acknowledged in the literature \cite{plummer2017enhancing} and form the motivation behind our proposed evaluation measures. However, our measures are different in the sense that a summary is not converted to text domain before evaluating. Rather, how important its selections are, or how diverse its selections are, is computed from the rich textual annotations in VISIOCITY. This is similar in spirit to \cite{sharghi2018improving}, but there it has been done only for egocentric videos. As noted by \cite{otani2019rethinking} "the limited number of evaluation videos and annotations further magnify this ambiguity problem". Our VISIOCITY framework precisely hits the nail by not only offering a larger dataset but also in proposing a richer evaluation framework better equipped at dealing with this ambiguity.

\section{VISIOCITY: A New Benchmark}
\label{sec3:visiocity}

We introduce VISIOCITY, a new challenging benchmark dataset for video summarization. VISIOCITY stands for Video Summarization based on Continuity, Intent and Diversity. As noted earlier, VISIOCITY is a large challenging dataset, which is an order of magnitude larger (both in terms of total number of hours as well as the duration of each video) compared to existing datasets for video summarization.

\subsection{Videos}
\label{subsec3:videos}
VISIOCITY is a diverse collection of 67 videos spanning across six different categories: TV shows (\emph{Friends}) , sports (Soccer), surveillance, education (Tech-Talks), birthday videos and wedding videos. The videos have an average duration of about 50 mins. VISIOCITY is compared to existing datasets in Table 1 and Summary statistics for VISIOCITY are presented in Table 2. Publicly available Soccer, Friends, Techtalk, Birthday and Wedding videos were downloaded from internet. TV shows contains videos from a popular TV series \emph{Friends}. They are typically more aesthetic in nature and professionally shot and edited. In sports category, VISIOCITY contains Soccer videos. These videos typically have well defined events of interest like goals or penalty kicks and are very similar to each other in terms of the visual features. Under surveillance category, VISIOCITY covers diverse settings like indoor, outdoor, classroom, office and lobby. The videos were recorded using our own surveillance cameras. These videos are in general very long and are mostly from static continuously recording cameras. Under educational category, VISIOCITY has tech talk videos with static views or inset views or dynamic views. In personal videos category, VISIOCITY has birthdays and wedding videos. These videos are typically long and unedited.

\begin{table}
\small{
\begin{center}
\begin{tabular}{|l|c|c|c|c|}
\hline
Domain & \# Videos & Duration &  Total Duration \\
\hline\hline
Sports(Soccer) & 12 & (37,122,\textbf{64}) & 12.8  \\
TVShows (Friends) & 12 & (22,26,\textbf{24}) & 4.8 \\
Surveillance & 12 & (22,63,\textbf{53}) & 10.6 \\
Educational & 11 & (15,122,\textbf{67}) & 12.28 \\ 
Birthday & 10 & (20,46,\textbf{30}) & 5 \\
Wedding & 10 & (40,68,\textbf{55}) & 9.2 \\ \hline
All & 67 & (26,75,\textbf{49}) & 54.68 \\
\hline
\end{tabular}
\label{tab1:ourdatasetstats}
\end{center}
\caption{Key Statistics of VISIOCITY. Third Column is (Min/Max/Avg) in Minutes and Fourth Column is in Hours.}}
\end{table}

\subsection{Annotations} \label{subsec3:annotations}
The annotations are in the form of concepts marked for each unit of annotation as against asking annotators to prepare actual summaries. Indirect ground truth offers several advantages. Firstly, being at a higher level it can be seen as a 'generator' of ground truth summaries and thus allows for multiple solutions (reference summaries) of different lengths with different desired characteristics and is easy to scale. Secondly, it is more informative. And thirdly, it makes the annotation process more objective and easier than asking the users to directly produce reference ground truth summaries.

As far as unit of annotation is concerned,our motivation was to keep the unit of annotation small enough to not contain redundancy within it and large enough to make the annotation process viable. Thus VISIOCITY has annotations at the level of shots (followed by a post processing step to merge very small shots and split big shots). Wherever the output of the shot detector was not satisfactory (for example in surveillance videos coming from static cameras), or in Soccer videos where visual content change is too frequent in spite of no change in semantic information,  we used fixed length (2-5 seconds) segments for annotation.  

The concepts are organized in categories instead of a long flat list. Example categories include 'actor', 'entity', 'action', 'scene', 'number-of-people', etc. The concept keywords within each category are carefully selected based on the category of the video through domain expertise. Categories provide a natural structuring to make the annotation process easier and also support for at least one level hierarchy of concepts for concept-driven summarization. 

In addition to the concept annotations for every shot, there are additional annotations to mark consecutive shots which \emph{together} constitute a cohesive event (we call such occurences as \emph{mega-events}). To better understand the idea of \emph{mega-events}, consider the case of a 'goal' event in Soccer. A model trained to learn importance scores (only) would do well to pick up the 'goal' snippet. However, such a summary will not be pleasing at all because what is required in a summary in this case is not just the ball entering the goal post but the build up to this event and probably a few shots as a followup. In this way, \emph{mega-event} annotations capture the notion of continuity.

\begin{figure}
    \centering
    \includegraphics[width=0.8\textwidth]{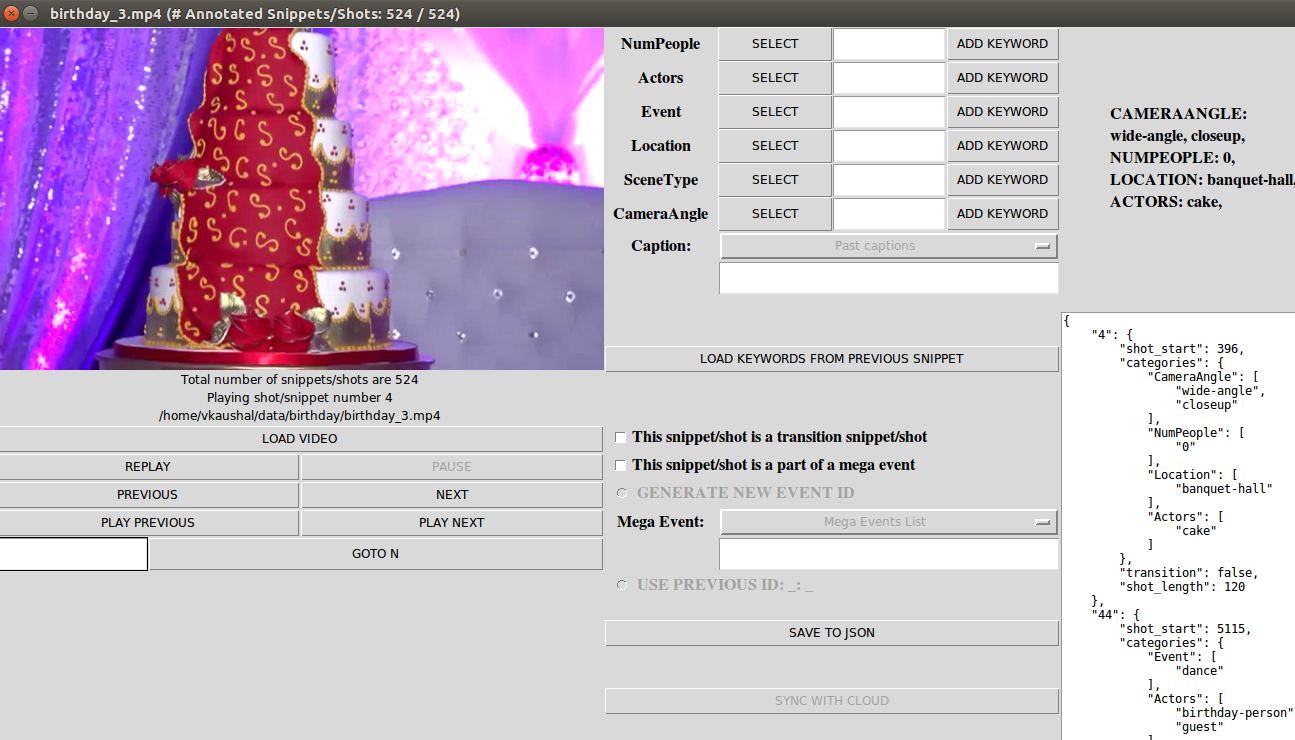}
    \caption{Annotation and visualization tool developed by us used in VISIOCITY framework}
    \label{fig:tool}
\end{figure}

While past work has made use of other forms of indirect ground truth like asking annotators to give a score or a rating to each shot \cite{song2015tvsum}, using textual concept annotations offers several advantages. First, it is easier and more accurate for annotators to mark all keywords applicable to a shot/snippet than for them to tax their brain and give a rating (especially when it is quite subjective and requires going back and forth over the video for considering what is {\it more important} or {\it less important}). Second, when annotators are asked to provide ratings, they often suffer from chronological bias. One work addresses this by showing the snippets to the annotators in random order~\cite{conf/cvpr/SongVSJ15} but it doesn't work for long videos because an annotator cannot remember all of these to be able to decide the relative importance of each. Third, the semantic content of a shot is better captured through text \cite{yeung2014videoset,plummer2017enhancing}. This is relevant from an 'importance' perspective as well as 'diversity' perspective. As noted earlier, two shots may look visually different but could be semantically same and vice versa. Text captures the right level of semantics desired by video summarization. Also, when two shots have same rating, it is not clear if they are semantically same or they are semantically different but equally important. Textual annotations brings out such similarities and dissimilarities more effectively. Fourth, as already noted, textual annotations make it easy to adapt VISIOCITY to a wide variety of problems. 
\paragraph{Annotation Protocol: } A group of 13 professional annotators were tasked to annotate videos (without listening to the audio) by marking all applicable keywords on a snippet/shot through a python GUI application developed by us for this task. It allows an annotator to go over the video unit by unit (shot/snippet) and select the applicable keywords using a simple and intuitive GUI (Figure~\ref{fig:tool}). It provides convenience features like copying the annotation from previous snippet, which comes in handy where there are are a lot of consecutive identical shots, for example in surveillance videos. 

Special caution was exercised to ensure high quality annotations. Specifically, the guidelines and protocols were made as objective as possible, the annotators were trained through sample annotation tasks, and the annotation round was followed by two verification rounds where both precision (how accurate the annotations were) and recall (whether all events of interest and continuity information has been captured in the annotations) were verified by another set of annotators. Whatever inconsistencies or inaccuracies were found and could be automatically detected, were included in our automatic sanity checks which were run on all annotations. 

\section{Ground Truth Summaries and Evaluation Framework} \label{sec4:evalframework}

\subsection{What characterizes a good summary?}
\label{subsec4:goodsumm}

\indent \textbf{Diversity} - A summary which does good on diversity is non redundant. It contains segments quite \emph{different} from one another. \emph{different} could mean different things in terms of content alone (i.e. one doesn't want two similar looking snippets in a summary) or in terms of content and time (i.e. one doesn't want visually similar \emph{consecutive} snippets, but \emph{does} want visually similar snippets that are separated in time) or in terms of the concepts covered (one does not want too many snippets covering the same concept and would rather want a few of all concepts). In surveillance videos for example, one would like to have a summary which  doesn't  have  too many visually  similar  consecutive and hence redundant snippets,  but  does  have  visually  similar  snippets  that  are separated  in  time. For instance, consider a video showing a person entering her office at three different times of the day. Though all three look similar (and will have identical concept annotations as well), all are desired in the summary. With regards to the quantitative formulation, we define the first flavor of diversity as $$Div_{sim}(X) = \max \min_{i, j \in X} d_{ij}$$ where $X$ is a subset of snippets. $d_{ij}$ is IOU measure between snippets $i$ and $j$ based on their concept vectors. For the other two flavors of diversity, we define diversity clustered: $$Div(X) = \sum_{i=1}^{|C|} \max_{j \in X \cap C_i} r_j$$ where $C$ are the clusters, which can be defined over time (all consecutive similar snippets form a cluster) or concepts (all snippets covering a concept belong to a cluster) and $r_j$ is the importance rating of a snippet $j$. When optimized, this function leads to the selection of the best snippet from each cluster. This can be easily extended to select a finite number of snippets from each cluster instead of the best one. 

\textbf{MegaEventContinuity} -element of continuity makes a summary pleasurable to watch. There is a thin line between modelling redundancy and continuity when it comes to visual cues of frames. Some snippets might be redundant but are important to include in the summary from a continuity perspective. To model the continuity, VISIOCITY has the notion of mega-events as defined above. To ensure no redundancy within a mega event, we define mega-events to be as tight as possible, meaning they should contain bare minimum snippets just enough to indicate the event. A non-mega event snippet is continuous enough to exist in the summary on its own and a mega event snippet needs other adjacent snippets to be included in the summary for visual continuity. We define this aspect quantitatively as follows 
$$MegaCont(X) = \sum_{i=1}^E r^{mega}(M_i) {|X \cap M_i|^2}$$ where, $E$ is the number of mega events in the video annotation, $r^{mega}(M_i)$ is the rating of the mega event $M_i$ and is equal to $\max_{\forall s \in M_i} r(s)$, $A$ is the annotation of video $V$, that is, a set of snippets such that each snippet $s$ has a set of keywords $K^{s}$ and information about mega event, $M$ is a set of all mega events such that each mega event $M_i$ ($i \in 1, 2, \cdots E$) is a set of snippets that constitute the mega event $M_i$

\textbf{Importance / Interestingness} - This is the most obvious characteristic of a good summary. For some domains like sports, there is a distinct importance of some snippets (for eg. score changing events) over other snippets. This however is not applicable for some other domains like tech talks where there are few or no distinctly important events. With respect to the annotations available in VISIOCITY, importance of a shot or snippet is defined by the ratings of the keywords of a snippet. These ratings come from a mapping function which maps keywords to ratings for a domain. The ratings are defined from 0 to 10 with 10 rated keyword being the most important and 0 indicated an undesirable snippet. We assign ratings to keywords based on their importance to the domain and average frequency of occurence. Given the ratings of each keyword, rating of a shot is defined as $r_s = 0$ if $\exists i: r_{K^s_i}=0$, and $r_s = \max_i r_{K^s_i}$ otherwise. Here $K^s$ is the set of keywords of a snippet $s$ and $r_{K^s_i}$ is the rating of a particular keyword $K^s_i$. Thus importance function can be defined as: $\mbox{Imp}(X) = \sum_{s \in X \cap A \setminus M}r(s)$. Note that when both importance and mega-event-continuity is measured, we define the importance only on the snippets which are not mega-events since the mega-event-continuity term above already handles the importance of mega-events.

\subsection{Recipe for automatic generation of ground truth summaries}
\label{subsec4:autosumm}
The indirect ground truth annotations in VISIOCITY allows us to generate ground truth summaries as follows.

The above functions modeling different characteristics of a good summary are a natural choice to be used as the building blocks of an ideal summary generator. A good summary should have these different characteristics to different degrees (and that is what makes on good summary different from another good summary, generating multiple correct answers). Thus we define a scoring function as a mixture of above components. This composite scoring function takes an annotated video {\it keywords} and {\it mega-events} defined over snippets/shots) and generates a set of candidate ({\it ground-truth}) summaries which supervised or semi-supervised summarization algorithms can use. 

Given $X$, a set of snippets of a video $V$, let $score(X)$ be defined as:
$$
score(X, \Lambda) = \lambda_1 \mbox{MegaCont}(X) + \lambda_2 \mbox{Imp}(X) + \lambda_3 \mbox{Div}(X)
$$
This scoring function is parametrized on $\Lambda$'s which assign the relative importance of these terms. This scoring function is approximately optimized via a greedy algorithm~\cite{minoux1978accelerated} to arrive at the candidate ground truth summaries.

Different configuration of $\lambda$ generates different summaries. We observe (as demonstrated in section~\ref{sec6:expresults}) that maximizing one characteristic doesn't necessarily and automatically lead to the maximization of another characteristic. In this sense these are orthogonal characteristics modelling different good aspects of a summary. Hence, some combinations would fare well on all while some would fare well on some, not so well on others. To understand what \emph{are} the best combinations corresponding to the multiple right answers, we borrow from the notion of Pareto optimality. Pareto optimality is a situation that cannot be modified so as to make any one individual or preference criterion better off without making at least one individual or preference criterion worse off. Beginning with a random element in the pareto-optimal set, we iterate over remaining elements to decide whether a new element should be added or old should be removed or new element should be discarded. This is decided on the basis of the performance on various measures. A configuration is better than another when it is better on all measures, otherwise it is not.

We show in section~\ref{sec6:expresults} that the automatic ground truth summaries so generated are at par with the human summaries both qualitatively and quantitatively.

There could be an alternate modeling of the hunt for good configurations. Consider each configuration as an allocation to some agents (the different scoring terms) such that an allocation yields different performance on different measures (the value seen by the agents). This setting allows us to use the notion of fair public decision making applying Nash social welfare equation, where the best allocation is one which is proportionally fair to all agents. In our experiments of comparing the automatic ground truth summaries so generated with human summaries, we found these to be similar in performance to the Pareto-optimal set defined above. In our experiments for comparing performance of different models on VISIOCITY, we have used the Pareto-optimal set of configurations to generate the automatic ground truth summaries. Our reason for this choice was motivated by the fact that while Pareto optimality gives multiple good configurations, proportional fairness gives only one. This however is not against the notion of multiple automatic summaries though, as even one configuration can potentially lead to many summaries by generating summaries "around" it.

For assessing a candidate summary we advocate against the use of a single measure. Since different measures model different characteristics, true and wholesome assessment of a candidate summary can only be done when all measures (including the existing measures like F score) are used. Results and observations from our extensive experiments coroborate this fact.

\section{A recipe to enhance an existing model}
\label{sec5:propmodel}
Following \cite{gygli2015video} we formulate the problem of automatic video summarization as a subset selection problem where a weighted mixture of set functions is maximized to produce an optimal summary for a desired budget. In our experiments we generate summaries in the range of 1\% to 5\% of the video length. The weights of the model are learnt using the following large margin framework as described in \cite{gygli2015video}. We train a simple model having only a submodular facility location term and a modular importance term with many automatic ground truth summaries and a margin loss which combines the feedback from different evaluation measures. The facility location function is defined as  $$f_{fl}(X) = \sum_{v \in V} \max_{x \in X} sim(v,x)$$ where $v$ is an element from the ground set $V$ and $sim(v, x)$ measures the similarity between element v and element x. Facility Location governs the representativeness aspect of the candidate summaries and is monotone submodular.\looseness-1. The importance scores are taken from the VASNet model \cite{fajtl2018summarizing} and the vsLSTM model \cite{zhang2016summary} trained on VISIOICTY. We call this proposed method VISIOCITY-SUM. We demonstrate that a simple model like this out performs the current techniques (state of the art on TVSum and SumMe) on VISIOCITY dataset because of learning from multiple ground truth summaries and learning from mutliple loss functions, better equipped at capturing different characteristics of a summary.

\section{Experiments and Results}
\label{sec6:expresults}

\subsection{Experiment Setup and Implementation Details}

For analysis of and comparison with human summaries, F1 score of any candidate summary is computed with respect to the human ground truth summaries following \cite{zhang2016summary}. We report both avg and max. To calculate F1 scores of human summaries with respect to human summary, we compute max and avg in a leave-one-out fashion. 

For analysis of and comparison of different techniques on the VISIOCITY dataset, we report their F1 scores computed against the automatically generated summaries as a proxy for human summaries. We generate 100 automatic summaries for each video. All target summaries are generated such that their lengths are 1\% to 5\% of the video length. We test the performance of three different representative state of the art techniques on the VISIOCITY benchmark  vsLSTM \cite{zhang2016summary} is a supervised technique that uses BiLSTM to learn the variable length context in predicting important scores. It learns from a combined ground truth in terms of aggregated scores. VASNet \cite{fajtl2018summarizing} is a supervised technique based on a simple attention based network without computationally intensive LSTMs and BiLSTMs. It learns from a combined ground truth in terms of aggregated scores and outputs a predicted score for each frame in the video. DR-DSN \cite{zhou2018deep} is an unsupervised deep-reinforcement learning based model which learns from a combined diversity and representativeness reward on scores predicted by a BiLSTM decoder. It outputs predicted score for every frame of a video. To generate a candidate machine generated summary from the importance scores predicted by vsLSTM, VASNET and DR-DSN, we follow \cite{zhang2016summary} to convert them into machine generated summary of desired length (max 5\% of original video). Our proposed model, VISIOCITY-SUM learns from multiple ground truth summaries and outputs a machine generated summary as a subset of shots.

In all tables, AF1 refers to Avg F1 score, MF1 refers to Max F1 score (nearest neighbor score), IMP, MC, DT, DC and DSi refer to the importance score, mega-event continuity score, diversity-time score, diversity-concept score and diversity-similarity score respectively, as measured by the functions introduced above. All figures are in percentages. 

\subsection{Degree of consistency and inconsistency in human summaries}

We asked a set of 11 users (different from the annotators) to generate human summaries for two randomly sampled videos of each domain. The users were supposed to look at the video and mark segments they feel should be included in the summary such that the length of the summary remains between 1\% to 5\% of the original video. The procedure followed was similar to that of SumMe~\cite{gygli2014creating}.

On closely examining the human selections, we observed that they are characterized by consistency to the extent there are important scenes in the video, for example a goal in Soccer videos. On the other hand, in the absence of such clear interesting events, the interestingness is more spread out and is more subjective, leading to higher inconsistency. We present more elaborate results and analysis in the Supplementray material. 

\subsection{Orthogonal characteristics of good summaries and our measures are good in capturing in those characteristics}

We assessed the human summaries using the set of performance measures presented above and found that while our measures get good scores on the human summaries, thus ascertaining their utility, there is also a lot of variation among the scores of different human summaries and this is captured well by our measures. We present the performance numbers in Tables 3-8. We also observe that a summary could be low on one measure, and yet is a good summary and this is captured by some other measure. Further, we also observe that the expected characteristics differ across different domains and the type of a particular video. Further, it is important to note that optimizing for one measure doesn't necessarily help the other measures thereby ascertaining that these measures are individually important. In the supplementary material we present a more detailed depiction of this interplay among different measures. As a simple example, a summary maximizing importance will do well to capture the goals in a soccer video, but unless there is an element of mega-event continuity in it, some shots preceeding the goal and following the goal will not be in the summary and the summary will not be visually pleasing. 

\subsection{Automatic summaries are at par with human summaries}

\begin{table}[ht]
\begin{center}
\begin{tabular}{|l|r|r|r|r|r|r|r|} 
\hline
Technique  & AF1 & MF1 & IMP & MC & DT & DC & DSi \\ \hline
Human & 30 & 45 & 56 & 55 & 75 & 84 & 85 \\
Uniform & 6 & 9 & 30 & 19 & 30 & 52 & 82 \\
Random & 5 & 9 & 30 & 22 & 30 & 51 & 81 \\
Auto & 27 & 37 & 83 & 88 & 82 & 90 & 80 \\   \hline
\end{tabular}
\end{center}
\caption{Performance of auto summaries for Soccer videos}
\end{table}

\begin{table}[ht]
\begin{center}
\begin{tabular}{|l|r|r|r|r|r|r|r|} 
\hline
Technique  & AF1 & MF1 & IMP & MC & DT & DC & DSi \\ \hline
Human & 24 & 38 & 55 & 46 & 72 & 70 & 88 \\
Uniform & 5 & 9 & 31 & 7 & 89 & 66 & 90 \\
Random & 6 & 13 & 31 & 16 & 28 & 38 & 85 \\
Auto & 25 & 41 & 87 & 69 & 76 & 85 & 81 \\   \hline
\end{tabular}
\end{center}
\caption{Performance of auto summaries for Friends videos}
\end{table}

\begin{table}[ht]
\begin{center}
\begin{tabular}{|l|r|r|r|r|r|r|r|} 
\hline
Technique  & AF1 & MF1 & IMP & MC & DT & DC & DSi \\ \hline
Human & 35 & 56 & 58 & 65 & 45 & 79 & 80 \\
Uniform & 6 & 8 & 12 & 9 & 12 & 49 & 55 \\
Random & 6 & 8 & 13 & 12 & 13 & 46 & 55 \\
Auto & 31 & 40 & 80 & 85 & 82 & 99 & 88 \\   \hline
\end{tabular}
\end{center}
\caption{Performance of auto summaries for Surveillance videos}
\end{table}

\begin{table}[ht]
\begin{center}
\begin{tabular}{|l|r|r|r|r|r|r|} 
\hline
Technique  & AF1 & MF1 & IMP & DT & DC & DSi \\ \hline
Human & 20 & 43 & 55 & 52 & 91 & 67 \\
Uniform & 7 & 9 & 49 & 29 & 45 & 60 \\
Random & 6 & 10 & 51 & 32 & 49 & 56 \\
Auto & 25 & 43 & 86 & 78 & 93 & 96 \\   \hline
\end{tabular}
\end{center}
\caption{Performance of auto summaries for Tech-Talk videos}
\end{table}

\begin{table}[ht]
\begin{center}
\begin{tabular}{|l|r|r|r|r|r|r|r|} 
\hline
Technique  & AF1 & MF1 & IMP & MC & DT & DC & DSi \\ \hline
Human & 21 & 31 & 56 & 38 & 48 & 70 & 83 \\
Uniform & 6 & 9 & 48 & 12 & 79 & 73 & 82 \\
Random & 6 & 10 & 48 & 16 & 47 & 57 & 78 \\
Auto & 17 & 30 & 86 & 81 & 63 & 91 & 84 \\   \hline
\end{tabular}
\end{center}
\caption{Performance of auto summaries for Birthday videos}
\end{table}

\begin{table}[ht]
\begin{center}
\begin{tabular}{|l|r|r|r|r|r|r|r|} 
\hline
Technique  & AF1 & MF1 & IMP & MC & DT & DC & DSi \\ \hline
Human & 21 & 39 & 57 & 39 & 46 & 69 & 76 \\
Uniform & 5 & 8 & 42 & 11 & 87 & 73 & 80 \\
Random & 5 & 9 & 42 & 18 & 40 & 54 & 78 \\
Auto & 14 & 21 & 81 & 79 & 71 & 95 & 88 \\   \hline
\end{tabular}
\end{center}
\caption{Performance of auto summaries for Wedding videos}
\end{table}

To compare how the automatically generated ground truth summaries fare with respect to the actual human summaries, we generated 100 automatic summaries per video of about the same length as the human summaries. We then compute scores of different measures and report the numbers in Tables 3-8.

We see that automatic and human summaries are both much better than random on all the evaluation criteria. Next, we see that both the human and the automatic summaries are close to each other in terms of the F1 metric. We also see that the automatic summaries have the highest Importance, Continuity and Diversity scores. This is not surprising as they are obtained at the first place by optimizing a combination of these criteria.

We also compare the human and automatic summaries qualitatively and see that a) it is very hard to distinguish the automatic summaries from humans though they are automatically generated, and b) they form very good visual summaries. We observe, though, that a perfect match is neither possible nor expected in the spirit of multiple right answers. We compare their selections in the supplementary material. 

\subsection{Analysis of performance of different techniques on VISIOCITY}

\begin{table}[h!t]
\begin{center}
\begin{tabular}{|l|r|r|r|r|r|r|r|} 
\hline
Technique  & AF1 & MF1 & IMP & MC & DT & DC & DSi \\ \hline
Auto                          & 59.3                      & 93.3                      & 83.2                    & 84.3                        & 82.6                         & 85.9                            & 76.2                        \\
DR-DSN                        & 2.8                       & 8.9                       & 23.7                    & 20.3                        & 23.2                         & 30.4                            & \textbf{83.4}                        \\
VASNET                        & 28.4                      & 43.4                      & \textbf{63}             & 49.3                        & \textbf{62.1}                & \textbf{67.4}                   & 75.2                        \\
vsLSTM                        & \textbf{31.9}             & \textbf{48.2}             & \textbf{62.2}           & \textbf{60.1}               & \textbf{62}                  & \textbf{69.5}                   & 76.5                        \\
VISIOCITY-SUM & \textbf{32.6} & \textbf{50.3} & \textbf{64.2} & \textbf{62.6} & \textbf{63.4} & \textbf{72.2} & 78.7 \\
Random & 3.4                       & 9.3                       & 25.7                    & 18.5                        & 25.5                         & 39.2                            & 80.5  \\ \hline           
\end{tabular}
\end{center}
\caption{Results on Soccer videos of VISIOCITY}
\end{table}

\begin{table}[h!t]
\begin{center}
\begin{tabular}{|l|r|r|r|r|r|r|r|r|} 
\hline
Technique  & AF1 & MF1 & IMP & MC & DT & DC & DSi \\ \hline
AUTO & 66.3 & 96.9 & 87.8 & 84.6 & 80.3 & 89.8 & 83.1 \\
DR-DSN & 4.3 & 9.4 & 19.1 & 6.9 & \textbf{65.7} & 51.5 & \textbf{98.5} \\
VASNET & \textbf{17} & \textbf{29.6} & \textbf{41} & \textbf{39.3} & 49 & \textbf{60.6} & 86.7 \\
vsLSTM & 15.5 & 27.2 & \textbf{40.4} & \textbf{39.2} & \textbf{64.7} & 59 & 91.1 \\
VISIOCITY-SUM & \textbf{17.4} & \textbf{31.2} & \textbf{42.5} & \textbf{40.5} & 50.2 & \textbf{64} & 90.3 \\
Random & 7.7 & 17.9 & 31.5 & 19.8 & 34.8 & 45.2 & 85.9 \\   \hline
\end{tabular}
\end{center}
\caption{Results on Friends videos of VISIOCITY}
\end{table}

\begin{table}[h!t]
\begin{center}
\begin{tabular}{|l|r|r|r|r|r|r|r|} 
\hline
Technique  & AF1 & MF1 & IMP & MC & DT & DC & DSi \\ \hline
Auto & 62.4 & 96.8 & 81.8 & 83.2 & 78.6 & 98 & 85.2   \\
DR-DSN & 10 & 17.7 & 33.6 & 20.2 & 21.8 & 54.5 & \textbf{57.2}   \\
VASNET & \textbf{19.4} & \textbf{31.4} & \textbf{39.5} & \textbf{42.6} & \textbf{28.4} & \textbf{65.4} & 37.6   \\
vsLSTM & 10.3 & 23.6 & 34.4 & 18.4 & 22.8 & 55.2 & \textbf{58.4}   \\
VISIOCITY-SUM & \textbf{20.5} & \textbf{32.6} & \textbf{41.7} & \textbf{44.3} & \textbf{29.6} & \textbf{68.2} & 38.5 \\
Random & 3.9 & 8 & 16.6 & 12 & 15.3 & 49.4 & 69.4   \\ \hline           
\end{tabular}
\end{center}
\caption{Results on Surveillance videos of VISIOCITY}
\end{table}

\begin{table}[h!t]
\begin{center}
\begin{tabular}{|l|r|r|r|r|r|r|r|} 
\hline
Technique  & AF1 & MF1 & IMP & DT & DC & DSi \\ \hline
Auto & 64.7 & 91.5 & 79.8 & 80.5 & 88.4 & 94 \\
DR-DSN & 13.5 & 22.5 & 49.3 & 24.8 & 29.9 & 35.2 \\
VASNET & \textbf{18.2} & \textbf{35.7} & 52.1 & \textbf{47.3} & \textbf{43.3} & \textbf{43.2} \\
vsLSTM & 15.1 & 32.2 & \textbf{60.3} & 38.8 & 35.3 & 41.7 \\
VISIOCITY-SUM & \textbf{18.7} & \textbf{37.5} & 53.2 & \textbf{50} & \textbf{45.8} & \textbf{45.5} \\
Random & 4.5 & 9.7 & 38.5 & 28 & 44 & 40.6 \\ \hline           
\end{tabular}
\end{center}
\caption{Results on TechTalk videos of VISIOCITY}
\end{table}

\begin{table}[h!t]
\begin{center}
\begin{tabular}{|l|r|r|r|r|r|r|r|} 
\hline
Technique  & AF1 & MF1 & IMP & MC & DT & DC & DSi \\ \hline
Auto & 67.3 & 97.2 & 89.7 & 88.6 & 68.1 & 90.8 & 81.3 \\
DR-DSN & 8.1 & 14.2 & 54.7 & 14.1 & \textbf{79.4} & 63.6 & \textbf{74.9} \\
VASNET & 21.6 & 37.6 & 50.1 & 30 & 36.2 & 47 & 48.7 \\
vsLSTM & \textbf{27.3} & \textbf{42.1} & \textbf{72.1} & \textbf{57.2} & 59.6 & \textbf{67.1} & 73.6 \\
VISIOCITY-SUM & \textbf{28} & \textbf{44.3} & \textbf{74.8} & \textbf{60.3} & \textbf{62} & \textbf{69.5} & \textbf{77.6} \\
Random & 6.9 & 14.2 & 51.8 & 16.9 & 49.2 & 54.8 & 70.3   \\ \hline           
\end{tabular}
\end{center}
\caption{Results on Birthday videos of VISIOCITY}
\end{table}

\begin{table}[h!t]
\begin{center}
\begin{tabular}{|l|r|r|r|r|r|r|r|} 
\hline
Technique  & AF1 & MF1 & IMP & MC & DT & DC & DSi \\ \hline
Auto & 55.4 & 94.4 & 83.9 & 74.7 & 67 & 88 & 85.7 \\
DR-DSN & 4.2 & 8.9 & 40.7 & 14.4 & \textbf{76.6} & \textbf{62} & \textbf{88.4} \\
VASNET & 4.5 & 14.4 & 46.5 & 22 & 44 & 52.7 & 84.9 \\
vsLSTM & \textbf{9} & \textbf{17.3} & \textbf{50.2} & \textbf{29.5} & 50.1 & 56.9 & 80.7 \\
VISIOCITY-SUM & \textbf{9.4} & \textbf{17.9} & \textbf{52.8} & \textbf{30.3} & \textbf{51.8} & \textbf{58.6} & 82.8 \\
Random & 3.5 & 10 & 41.1 & 16.3 & 40.6 & 51.6 & 80   \\ \hline           
\end{tabular}
\end{center}
\caption{Results on Wedding videos of VISIOCITY}
\end{table}

As another significant contribution, we present the performance of some representative summarization techniques tested on VISIOCITY benchmark and report their numbers in Tables 9-14. We make the following observations: a) DR-DSN tries to generate a summary which is diverse. As we can see in the results, it almost always gets high score on the diversity term. Please note that the way we have defined these diversity measures, diversity-concept (DC) and diversity-time (DT) have an element of importance in them also. On the other hand, diversity-sim (DSi) is a pure diversity term where DR-DSN almost always excels. b) Due to this nature of DR-DSN, when it comes to videos where the interestingness stands out and importance clearly plays a more important role, DR-DSN doesnt perform well. In such scenarios, vsLSTM is seen to perform better, closely followed by VASNET. c) It is also interesting to note that while two techniques may yield similar scores on one measure, for example vsLSTM and VASNET for Soccer videos (Table 9), one of them, in this case vsLSTM, does better on mega-event continuity and produces a desirable characteristic in the summary. This further strengthens our claim of having a set of measures evaluating a technique or a summary rather than over dependence on one, which may not fully capture all desirable characteristics of good summaries. d) We also note that even though DR-DSN is an unsupervised technique, it is a state of the art technique when tested on tiny datasets like TVSum or SumMe, but when it comes to a large dataset like VISIOCITY, with more challenging videos, it doesn't do well, especially on those domains where there are clearly identifiable important events for example in Soccer (goal, save, penalty etc.) and Birthday videos (cake cutting, etc.). In such cases, models like vsLSTM and VASNET perform better as they are geared towards learning importance. In contrast, since the interstingness level in videos like Surveillance and Friends is more spread out, DR-DSN does relatively well even without any supervision. 

We also report the performance of our simple model extension recipe VISIOCITY-SUM and compare them with the different models. We experimented with four flavors of our model. Three using the margin loss coming from importance, mega event continuity and diversity one at a time, and the fourth one learnt using a combined loss function. We found the fourth model to perform the best among these four and we have reported the numbers from this model (VISIOCITY-SUM) in the tables. A quick comparison with other models reveal that almost on all different characteristics, this new model manages to do well and better than all other techniques.

\section{Conclusion}

Identifying the need of the community, we presented VISIOCITY, a large benchmarking dataset and demonstrated its effectiveness in real world setting. To the best of our knowledge, it is the first of its kind in the scale, diversity and rich concept annotations. We introduce a recipe to automatically create ground truth summaries typically needed by the supervised techniques. We also extensively discuss and demonstrate the issue behind multiple right answers making the evaluation of video summaries a challenging task. Motivated by the fact that different good summaries have different characteristics and are not necessarily better or worse than the other, we propose an evaluation framework better geared at modeling human judgment through a suite of measures than having to overly depend on one measure. Finally through extensive and rigorous experiments we report the strengths and weaknesses of some representative state of the art techniques when tested on this new benchmark. Motivated by a fundamental problem in current supervised approaches, of learning from a single combined ground truth summary and/or learning from a single loss function tailored and optimizing \emph{one} characteristic, our attempt to make simple extension to an existing mixture model technique gives encouraging results. We hope our attempt to address the multiple issues currently surrounding video summarization as highlighted in this work, will help the community advance the state of the art in video summarization. 
\section*{Acknowledgements}

This work is supported in part by the Ekal Fellowship (www.ekal.org) and National Center of Excellence in Technology for Internal Security, IIT Bombay (NCETIS, https://rnd.iitb.ac.in/node/101506)

%%
%% The next two lines define the bibliography style to be used, and
%% the bibliography file.
\bibliographystyle{abbrv}
\bibliography{egb}

\end{document}